\documentclass{article}

\usepackage[square,comma,numbers,sort&compress]{natbib} % bibliography helpers

\usepackage[preprint]{neurips_2024}

\usepackage[utf8]{inputenc} % allow utf-8 input
\usepackage[T1]{fontenc}    % use 8-bit T1 fonts
\usepackage{hyperref}       % hyperlinks
\usepackage{url}            % simple URL typesetting
\usepackage{booktabs}       % professional-quality tables
\usepackage{amsfonts}       % blackboard math symbols
\usepackage{nicefrac}       % compact symbols for 1/2, etc.
\usepackage{microtype}      % microtypography
\usepackage{amsmath, amssymb}
\usepackage{graphicx}
\usepackage{xcolor}
\graphicspath{./figures}

\newcommand{\ut}{UTVI}
\newcommand{\mc}{MCVI}
\newcommand{\smp}{SMP}

\title{Few-sample Variational Inference of Bayesian Neural Networks
    with Arbitrary Nonlinearities}

\author{David J. Schodt \\
    Teledyne Scientific \& Imaging, LLC \\
    \texttt{david.schodt@teledyne.com}}

\begin{document}

    \maketitle

    \begin{abstract}
        Bayesian Neural Networks (BNNs) extend traditional neural networks to
        provide uncertainties associated with their outputs.  On
        the forward pass through a BNN, predictions (and their uncertainties)
        are
        made either by Monte Carlo sampling network weights from the learned
        posterior or by analytically propagating statistical moments through the
        network.  Though flexible, Monte Carlo sampling is
        computationally expensive and can be infeasible or impractical under
        resource constraints or for large networks.  While moment propagation
        can
        ameliorate the computational costs of BNN inference, it can be
        difficult or
        impossible for networks with arbitrary nonlinearities, thereby
        restricting
        the possible set of network layers permitted with such a scheme.  In
        this
        work, we demonstrate a simple yet effective approach for propagating
        statistical moments through arbitrary nonlinearities with only 3
        deterministic samples, enabling few-sample variational inference of BNNs
        without restricting the set of network layers used.  Furthermore, we
        leverage this approach to demonstrate a novel nonlinear activation
        function that we use to inject physics-informed prior information into
        output nodes of a BNN.
    \end{abstract}

    \section{Introduction}
    Bayesian Neural Networks (BNNs)
    \cite{mackay_practical_1992,neal_bayesian_1996}
    treat learnable parameters of a neural network as distributions,
    enabling predictive power and training insights generally not available from
    traditional neural networks.  Of particular interest is their ability to
    express calibrated uncertainties in their predictions, a feature that makes
    BNNs desirable in many (if not most) real-world applications.

    In the context of Bayesian inference, BNNs are trained to learn the
    posterior
    distribution of network weights given the observed data and prior
    distributions
    over the weights.  Since computing the
    true posterior distribution is generally intractable, approximate inference
    techniques, such as variational inference, are often used
    \cite{hinton_1993,barber_1998,graves_2011}.  With the variational
    approximation, BNN inference becomes practical and often relatively simple
    to
    implement, for instance using Bayes by backprop \cite{blundell_weight_2015}.
    Despite the relative ease of implementation, BNN inference can be
    prohibitively
    expensive computationally in some applications, such as when compute
    resources
    are limited or when the number of learnable weight distributions is
    especially
    large.

    Using the mean-field approximation (i.e., weights are conditionally
    independent
    given the data and hence the posterior is factorized as a product of the
    weight
    densities) and assuming Normally distributed weights, converting a standard
    neural network to a BNN at most doubles the number of learnable parameters
    (the
    mean and variance of each weight distribution) and hence is not particularly
    taxing computationally.  During inference, however, the
    need for Monte Carlo sampling can require 10s or 100s of samples to
    accurately
    characterize the output distribution, particularly for regression tasks.
    Alternatively, the need for Monte Carlo sampling can be eliminated or
    greatly
    reduced by propagating statistical moments (e.g., mean and variance) through
    the network, providing a deterministic prediction with reduced
    computational complexity \cite{wu_deterministic_2019,
        hausmann_sampling-free_2020,schodt_2024}.
    Unfortunately, existing moment propagation-based inference techniques only
    work
    on a restricted set of network layers, with addition of new layers requiring
    analytic derivation or approximation of associated propagation rules (see,
    for
    example, the equations needed to propagate a Normally distributed random
    variable through the popular leaky-ReLU activation presented in
    \cite{schodt_2024}) or yet again relying on Monte Carlo sampling.  The lack
    of
    extensibility to novel nonlinearities reduces the potential impact of moment
    propagation-based BNN inference.

    In this work, we present an effective and relatively straightforward method
    to
    enable moment propagation through BNNs with arbitrary nonlinearities.  In
    particular, we show that the unscented transform \cite{julier1997}
    (popularized
    in Kalman
    filtering applications) enables few-sample deterministic BNN inference with
    significantly
    reduced computational costs compared to Monte Carlo sampling without
    sacrificing
    performance.  We demonstrate the utility of the unscented transform for BNN
    inference on a simple multi-layer perceptron with known analytic propagation
    rules and on a more complex convolutional neural network with a novel,
    application-specific activation function for which analytic moment
    propagation
    rules have not been derived.

    \section{Background}
    \label{sec:background}
    \subsection{Variational inference of Bayesian neural networks}
    \label{sec:variational_inference}
    Our terminal objective for BNN inference is to estimate a predictive
    distribution that describes the probability of observing an output (denoted
    by
    $\mathbf{y}$) given an input (denoted by $\mathbf{x}$) conditioned on the
    previous observations.  For instance, we may
    wish to estimate the position of an object (output) within an image
    (input), in
    which case we seek a distribution over possible positions conditioned on the
    image.  We denote our BNN by the distribution $p(\mathbf{y} | \mathbf{x},
    \mathbf{w})$ parameterized by weights $\mathbf{w}$.  The training data is
    denoted as $\mathcal{D} \equiv\{(x_i, y_i)\}_{i=1}^m$.  The desired
    predictive
    distribution is thus
    \begin{align}
        \label{eqn:predictive_distribution}
        p(\mathbf{y} | \mathbf{x}, \mathcal{D})
        &= \mathbb{E}_{p(\mathbf{w} | \mathcal{D})} [p(\mathbf{y} |
        \mathbf{x}, \mathcal{D}, \mathbf{w})] \nonumber \\
        &= \int d\mathbf{w} \, p(\mathbf{y}
        |
        \mathbf{x}, \mathcal{D}, \mathbf{w}) p(\mathbf{w} | \mathcal{D})
    \end{align}
    The variational approximation involves replacing the (generally intractable)
    distribution $p(\mathbf{w} | \mathcal{D})$ by a variational distribution
    $q(\mathbf{w} | \mathbf{\theta})$ parameterized by learnable parameters
    $\theta$.

    \subsubsection{Monte Carlo sampling}
    Although the variational approximation simplifies BNN inference, computing
    the
    integral in Eqn. \ref{eqn:predictive_distribution} remains intractable in
    general.  As such, Monte Carlo sampling is often employed to approximate the
    integral \cite{neal_bayesian_1996,blundell_weight_2015}.  This is achieved
    by
    taking $n$ samples from the variational
    posterior $q(\mathbf{w} | \mathbf{\theta})$ (which, by design, we can sample
    from) and computing the associated $n$ forward passes
    $\{\mathbf{y}_j\}_{j=1}^n$ through
    the BNN.  Moments
    of the predictive distribution are then computed as needed, for example the
    expected value and variance of $\mathbf{y}$
    \begin{align*}
        \mathbb{E}[\mathbf{y}] &= \int d\mathbf{y} \, \mathbf{y} p(\mathbf{y}
        | \mathbf{x}, \mathcal{D})
        \approx \frac{1}{n} \sum_{j=1}^{n} \mathbf{y}_j = \bar{\mathbf{y}}\\
        \mathbb{V}[\mathbf{y}] &\approx \frac{1}{n-1} \sum_{j=1}^{n}
        (\mathbf{y}_j-\bar{\mathbf{y}})^2
    \end{align*}

    \subsubsection{Moment propagation}
    Monte Carlo integration of Eqn. \ref{eqn:predictive_distribution} is
    computationally expensive and in some use cases may not be practical.  As
    an example, consider using a BNN to solve a 1-dimensional regression task
    where the data are Normally distributed as $y|x \sim \mathcal{N}(\mu(x),
    \sigma(x)^2)$.  The standard error of the posterior predictive mean
    $\hat{\bar{y}}$ computed from $n$ independent Monte Carlo samples of the BNN
    would be $\mathrm{se}(\hat{\bar{y}}) = \sigma / \sqrt{n}$,
    hence estimating the posterior predictive mean with standard error of
    $0.1\sigma$ would require at least 100 samples.  Such a large
    number of samples is not practical when compute resources are limited or for
    large networks.

    The above example motivates a more computationally efficient inference
    approach.
    Thankfully, under the mean-field approximation that neurons are
    conditionally
    independent, it is possible to analytically propagate the mean and
    variance through many common network layers (e.g., fully-connected,
    convolution, average pool) by taking advantage of simple properties of mean
    and variance of independent random variables:
    \begin{align}
        \label{eqn:expectation_rules}
        \mathbb{E}[aX + b] &= a \mathbb{E}[X] + b \nonumber \\
        \mathbb{V}[aX + b] &= a^2 \mathbb{V}[X] \nonumber \\
        \mathbb{E}[XY] &= \mathbb{E}[X] \mathbb{E}[Y]
        \nonumber \\
        \mathbb{V}[XY] &= \mathbb{V}[X] \mathbb{V}[Y] + \mathbb{V}[X]
        \mathbb{E}[Y]^2 + \mathbb{E}[X]^2 \mathbb{V}[Y]
    \end{align}
    These properties were used in
    \cite{wu_deterministic_2019,hausmann_sampling-free_2020,schodt_2024} to
    eliminate the need for Monte Carlo sampling.  For nonlinear layers, we must
    appeal to additional assumptions and more complicated computations, e.g.,
    for
    ReLU \cite{hausmann_sampling-free_2020} or leaky-ReLU \cite{schodt_2024}.
    In
    practice, this restricts the set of layers available to us within these
    inference frameworks to those for which analytic expressions of their
    moments
    are available, or otherwise requires us to again use Monte Carlo sampling
    for
    those layers.

    \section{Few-sample variational inference}
    In this section, we describe a simple technique for propagating mean and
    variance through nonlinear network layers.  Our approach leverages the
    unscented transform popular in Kalman filtering, which allows us to
    approximate
    the effect of nonlinear network layers on an input distribution using only a
    few \emph{deterministic} samples.  Below, we describe the basics of the
    unscented transform and describe our implementation.

    \subsection{Unscented transform}
    To motivate the unscented transform, consider Normally distributed
    observations
    $x \sim \mathcal{N}(\mu, \sigma^2)$.  A nonlinear function $f$ transforms
    the
    observations as $f(x)$ and we wish to compute $\mathbb{E}[f(x)]$
    and $\mathbb{V}[f(x)]$.  One approach would be to linearize $f$,
    which would allow us to use the properties in Eqn.
    \ref{eqn:expectation_rules}
    to propagate mean
    and variance through $f$.  Alternatively, we could
    transform random samples $x$ with the true nonlinear function $f$ and
    compute
    descriptive statistics from the result, as done in Monte Carlo sampling.
    Linearization requires us to compute the Jacobian and thus precludes
    application to ``black box'' nonlinearities \cite{julier1997} while
    Monte Carlo sampling is computationally expensive, hence neither approach is
    particularly appealing.  This is the motivation for the unscented transform.

    Briefly, the unscented transform involves selecting a small number
    (typically
    at least $2d+1$ in $d$ dimensions) of deterministic ``sigma points'' in the
    domain of the
    nonlinearity, transforming the
    sigma points, and then computing descriptive statistics of the transformed
    sigma points.  Given sigma points $\{\chi_k\}_{k=1}^s$ and associated
    weights
    $\{\gamma_k\}_{k=1}^s$, the
    mean
    and variance of the distribution of $f(x)$ are
    approximated from the $s$ transformed sigma points as
    \begin{align}
        \mathbb{E}[f(x)] &\approx \sum_{k=1}^s \gamma_k f(\chi_k) =
        \bar{f} \label{eqn:ut_mean} \\
        \mathbb{V}[f(x)] &\approx \sum_{k=1}^s \gamma_k (f(\chi_k) -
        \bar{f}) ^ 2 \label{eqn:ut_var}
    \end{align}
    Sigma points and their associated weights are defined from the input
    distribution such that $\sum_{k=1}^s \gamma_k \chi_k = \mu$ and
    $\sum_{j=1}^m
    \gamma_k
    (\chi_k - \mu)^2 = \sigma^2$.

    \subsection{Our approach}
    \label{sec:utvi}
    To achieve few-sample variational inference of BNNs, we use the unscented
    transform to propagate the mean and variance through layers of our BNN.  We
    use
    this approach in conjunction with the analytical moment propagation method
    in
    \cite{schodt_2024}, for instance to propagate through layers where
    propagation
    rules are unknown or intractable\footnote{There is a caveat: as presented
    here,
        the unscented transform cannot be used to propagate through network
        layers that
        are both nonlinear \emph{and} stochastic, hence our ``Bayesian'' layers
        are
        restricted to operations through which we can propagate analytically.}.

    In this work, we use the mean-field approximation and assume the learnable
    parameters $\mathbf{w}$ of our BNNs are Normally distributed, hence our
    variational posterior takes the form $q(\mathbf{w} | \theta) = \prod_{j=1}^m
    q(w_j | \mu_j, \sigma_j^2)$ where $q$ is the density of the Normal
    distribution
    with parameters $\theta = \{(\mu_j, \sigma_j^2)\}_{j=1}^M$.  We additionally
    assume that outputs of each layer parameterized by $w_j$ is again Normally
    distributed.  We note that the
    unscented transform can be applied even when $q(\mathbf{w} |\theta)$ is not
    isotropic, however we do not explore that case here.  Furthermore, the
    recently
    developed generalized unscented transform \cite{ebeigbe2021generalized}
    could
    be used to eliminate the need to assume Normality.

    To propagate through the nonlinear layers of our network, we note that the
    mean-field approximation allows us to apply a one dimensional unscented
    transform independently for each operation parameterized by $w_j$.  For each
    input $x_i \sim \mathcal{N}(\mu_i, \sigma_i^2)$, we define sigma points and
    associated weights as in \cite{julier1997}:
    \begin{alignat}{3}
        \label{eqn:sigmas}
        \chi_{0,i} &= \mu_i, \quad &\gamma_{0,i} &= \frac{\kappa}{\kappa + 1}
        \nonumber \\
        \chi_{1,i} &= \mu_i - \sigma_i\sqrt{\kappa + 1}, \quad &\gamma_{1,i} &=
        \frac{1}{2(\kappa
            +
            1)} \nonumber \\
        \chi_{2,i} &= \mu_i + \sigma_i\sqrt{\kappa + 1}, \quad &\gamma_{2,i} &=
        \frac{1}{2(\kappa
            +
            1)}
    \end{alignat}
    where $\kappa$ is a parameter used to influence the spread of sigma
    points.  In
    this work, we use $\kappa=2$, though in principle allowing a learnable
    $\kappa>0$ may be beneficial.
    The mean and variance of $f(x_i)$ is then computed
    using Eqns. \ref{eqn:ut_mean} and \ref{eqn:ut_var} using the sigma points
    and
    weights in Eqn. \ref{eqn:sigmas}.  This process is repeated for all layer
    inputs $i$ and is then continued sequentially through the forward pass of
    the
    network, eventually yielding a mean and variance estimate for each output
    node
    of the BNN.

    \section{Experimental demonstration}
    In this section, we demonstrate our few-sample BNN inference approach on
    regression tasks solved by Bayesian fully-connected networks and Bayesian
    convolutional neural networks.  For each experiment, we compare our results
    to
    those obtained by Monte Carlo sampling moment propagation through network
    nonlinearities.  Unless otherwise noted, propagation through remaining
    network layers (e.g., linear layers) is performed using the method
    presented in \cite{schodt_2024}.  We use the following
    notation to indicate which method is used to propagate moments through
    network
    nonlinearites: ``\mc{}'' (Monte Carlo variational inference) for Monte Carlo
    sampling, ``\smp{}'' (simplified moment propagation) for purely
    analytical propagation as in \cite{schodt_2024}, and ``\ut{}'' (unscented
    transform variational inference) for the
    unscented transform method described in Section \ref{sec:utvi}.

    \subsection{Training details}
    \label{sec:training}
    All networks were trained using the AdamW optimizer
    \cite{loshchilov_decoupled_2019} with a learning rate $\alpha=0.001$,
    $\beta_1=0.9$, $\beta_2=0.999$, $\epsilon=10^{-8}$, and a weight decay
    $\lambda=0.01$.  We used Bayes by backprop \cite{blundell_weight_2015} with
    the
    Normal distribution likelihood for the likelihood term in the ELBO loss.  We
    scaled the KL-loss term of the ELBO at each epoch $l=1,2,\ldots,M$ by a
    multiplicative factor $\phi(l) = \frac{2^{M-l}}{2^M-1}$ as in
    \cite{schodt_2024}.  All results shown were averaged over 10 networks
    trained
    identically from randomly seeded initializations.  We trained all networks
    using purely synthetic data generated on-the-fly, so networks never saw the
    same data more than once during training.

    \subsection{Regression with a fully-connected BNN}
    \label{sec:fc_bnn_exp}
\begin{figure}[htbp!]
        \centering
        \includegraphics[width=\textwidth]{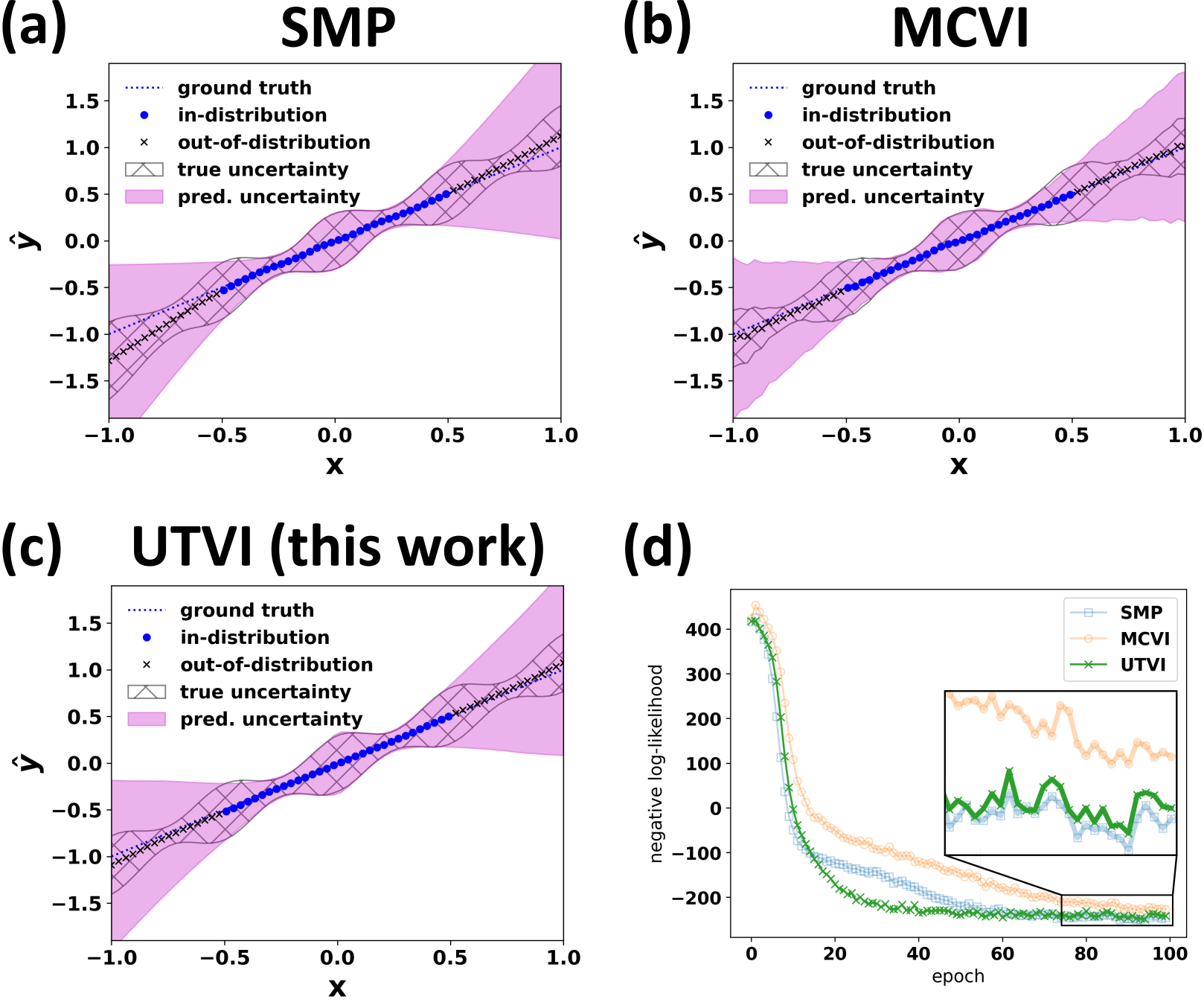}
        \caption{\textbf{\ut{} outperforms \mc{} and matches \smp{}.} BNN
            predictions using (a) \smp{}, (b) \mc{} with 3 samples, and (c)
            \ut{}.  All
            data points
            represent the average over 10 networks trained independently from
            distinct
            random seeds.  Note that \mc{} with 3 samples appears less noisy
            than
            expected due to the averaging over 10 networks (see Figure
            \ref{fig:mc_3samples} for an example output from a single
            network).  ``true
            uncertainty'' is defined as $\sqrt{\epsilon(x)^2 +
                \delta(x)^2}$ where $\epsilon(x)$ is the standard deviation of
                the
            simulated noise and
            $\delta(x)$ is the observed deviation from the prediction to the
            ground
            truth.  (d) Reconstruction loss after each epoch (negative
            log-likelihood of
            in-distribution validation given the trained model) averaged across
            each of
            the 10 networks.}
        \label{fig:fc_bnn}
    \end{figure}

    As an initial proof-of-concept, we use a fully-connected BNN with two hidden
    layers to model the function $y =
    x + \epsilon(x)$ where $\epsilon(x)
    \sim \mathcal{N}(0, \sigma(x)^2)$ and $\sigma(x)=0.1 + 0.2 \sin{(2 \pi x -
        \pi/2)}^2$.  Each hidden layer consists of 128 neurons and a leaky-ReLU
    activation.  This problem serves as a convenient baseline because, for
    Normally
    distributed data, we can analytically propagate through the network using
    the
    \smp{}
    method (i.e., we can analytically compute $p(\mathbf{y} | \mathbf{x},
    \mathbf{w})$).  For
    \mc{}, we propagate moments through the leaky-ReLU layers with $n=3$ Monte
    Carlo
    samples (for a direct comparison to the 3 samples used for \ut{}).  For
    \ut{}, we propagate moments through the leaky-ReLU layers with 3
    sigma points as described in Section \ref{sec:utvi}.  The results of this
    experiment are shown in Figure \ref{fig:fc_bnn}, wherein all results were
    averaged over 10 independently trained model seeds (see Figure
    \ref{fig:mc_3samples} for an example of single-model results).
    Qualitatively,
    \ut{}
    (Figure
    \ref{fig:fc_bnn}(c)) matches
    the performance of \smp{} (Figure \ref{fig:fc_bnn}(a)) while exceeding the
    performance of \mc{} (Figure \ref{fig:fc_bnn}(b)).  Interestingly, as seen
    in
    Figure \ref{fig:fc_bnn}(d), \ut{} seems to have better training
    characteristics
    as compared to \smp{}.  We posit that this is due to the reduced number of
    floating point operations used by \ut{} as compared to analytical moment
    propagation through the leaky-ReLU layers in the network (see
    \cite{schodt_2024} for analytical propagation rules), hence \ut{} may have
    more
    numerically stable gradients during training.

\begin{figure}[htbp!]
        \centering
        \includegraphics[width=\textwidth]{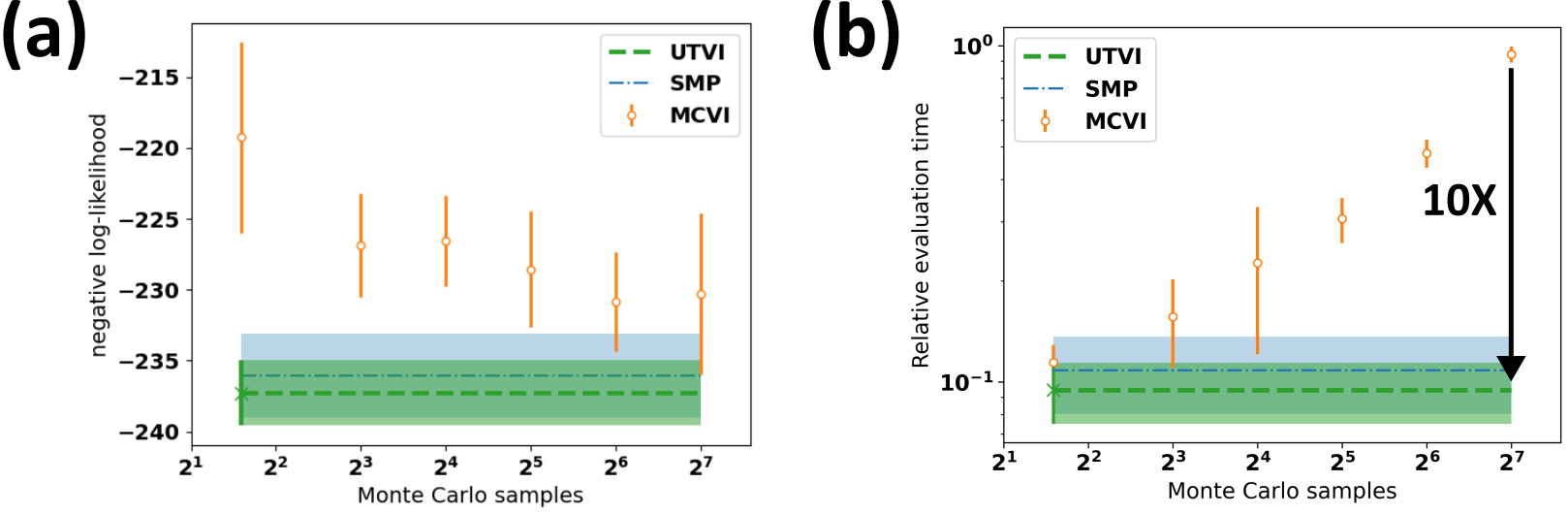}
        \caption{\textbf{\ut{} is 10X faster than \mc{} at similar level of
                performance.}  (a) Minimum negative log-likelihood
            achieved
            by the best models on the evaluation set for \mc{} models trained
            and
            evaluated with a varying
            number of samples.  All
            data points shown were averaged over 10 independently trained models
            initialized with different random seeds.  Error bars indicate $\pm$1
            standard deviation over the 10 models.  ``Best'' models were
            selected by
            taking the training checkpoint that achieved the lowest negative
            log-likelihood over the evaluation set.  Results from \ut{} with 3
            samples and \smp{} (sampling-free) are shown
            for comparison.  (b) Relative
            evaluation time
            averaged over 10 batches of 1024 network evaluations on an NVIDIA
            RTX
            A1000
            laptop GPU.  At
            $2^7$ samples (with which \mc{} achieves roughly the same
            performance as
            \ut{}), inference using \ut{} is approximately 10X faster than with
            \mc{}.}
        \label{fig:mc_samples}
    \end{figure}

    To emphasize the savings in computational costs achieved by \ut{}, we
    tracked
    the compute time needed for forward passes through each network on data
    batches
    of size 1024, with the number of samples used for \mc{} varying until
    reaching
    comparable performance (in terms of the minimum validation set negative
    log-likelihood achieved during training) to \ut{}.  The results of this
    study
    are shown in Figure
    \ref{fig:mc_samples}.  As seen in Figure \ref{fig:mc_samples}(a), \ut{}
    with 3
    sigma points exceeds the performance of \mc{} until roughly $2^7$ Monte
    Carlo samples are used.  Accordingly, as shown in Figure
    \ref{fig:mc_samples},
    \ut{} is approximately 10X more computationally efficient than \mc{} at a
    comparable level of performance.  Furthermore, we see that \ut{} matches
    the performance of \smp{} (Figure \ref{fig:mc_samples}(a)) with comparable
    computational costs (Figure \ref{fig:mc_samples}(b)).

    \subsection{Sub-pixel object localization}
    To demonstrate \ut{} on a more practically-relevant task, we use a
    convolutional
    BNN to quantify physical characteristics of an object within an image.
    Specifically, we use a BNN to estimate the two-dimensional position of a
    simulated fluorescent molecule imaged by a pixelated sensor, as well as the
    total number of photons incident on the sensor (a common task in, e.g.,
    localization microscopy
    \cite{lidke_2005,betzig_2006,rust_2006}, where predictive uncertainties are
    necessary for subsequent analyses).  For this task, we are
    particularly interested in injecting physics-informed constraints or other
    prior
    knowledge to restrict or otherwise guide the BNN towards a certain solution
    space.  For example, we may have prior knowledge that the object of
    interest is
    likely to be well-centered within the image, or that the total number of
    collected photons follows a known distribution.  We describe this
    experiment in
    more detail in the following sections.

    \subsubsection{Simulation}
    \label{sec:gauss_sim}
    To simulate data pairs $(\mathbf{x}_i, \mathbf{y}_i) = (\mathbf{x}_i,
    [\mathbf{r}_i, N_i(\mathbf{r}_i)])$, we sample a two-dimensional emitter
    position
    $\mathbf{r}_i \sim \mathcal{N}(\mathbf{0}, \sigma_\mathrm{r}^2 \mathbf{I})$
    with $\sigma_\mathrm{r} = 2$ pixels,
    place a
    Gaussian blob with standard deviation
    $\sigma_b=0.21\lambda/\mathrm{na}=1.05$
    pixels (the Gaussian point spread function standard deviation \cite{Zhang07}
    corresponding to a simulated
    emission wavelength of 6 pixels measured with numerical aperture
    $\mathrm{na}=1.2$)
    normalized to
    integrate to
    $N=100$ photons at $\mathbf{r}_i$ in a
    pixelated image, and then corrupt the image with Poisson noise to yield
    $\mathbf{x}_i$.  Note that our final image is a square with side length of
    $L=8$
    pixels, so the detected photon count $N_i \leq N$ decreases for emitters
    near
    the
    edge of the image.  Examples of simulated data are provided in Figure
    \ref{fig:ex_data}.

    \subsubsection{Injecting prior knowledge into BNN output nodes}
    \label{sec:icdf_activations}
    To incorporate prior knowledge about our data into the BNN, we implement
    custom
    nonlinear activation functions at the output nodes of our BNN.  We aim to
    mimic
    inverse transform sampling by using a (to the best of our knowledge) novel
    activation chain consisting of a scale function $f(x)$ that transforms
    inputs as $f: x \rightarrow [0, 1]$ followed by the inverse cumulative
    distribution $F^{-1}(f(x))$ of a prior distribution on the output node.
    Ideally, the scale function should satisfy $f(x) \sim \mathcal{U}(0, 1)$;
    in practice, we have observed the sigmoid function $f(x) = 1 /
    (1+\exp(-x/a))$ with learnable parameter $a$ to achieve our intended
    outcome.

    For the output nodes predicting $\mathbf{r}_i$, we choose the prior
    $\mathcal{N}(\mathbf{0}, \sigma^2 \mathbf{I})$ with $\sigma=2$, biasing
    predicted positions towards the center of the image.

    For the output node predicting the measured photon count, we choose the
    prior
    $\mathcal{N}(\mu=N_i(\hat{\mathbf{r}}_i),
    \sigma^2=N_i(\hat{\mathbf{r}}_i))$ to
    approximate the Poisson distribution with mean $N_i(\hat{\mathbf{r}}_i)$.
    This
    activation, which is applied \emph{after} computing position nodes, allows
    us
    to inject known physics into our network: the number of
    collected photons is (1) expected to be Poisson and (2) dependent on the
    position of the emitter within the restricted field of view captured by the
    image.  During inference, we estimate $N_i$ by integrating an isotropic
    Normal
    distribution with mean $\hat{\mathbf{r}}_i$ and variance $\sigma_b^2$
    over the image domain:
    \begin{align*}
        N_i(\hat{\mathbf{r}}_i) &= \frac{N}{2\pi \sigma_b^2} \int_{L
            \times L} d\mathbf{r} \, \exp{\left( -\frac{(\mathbf{r} -
                \hat{\mathbf{r}}_i)^2}{2 \sigma_b^2} \right)} \\
        &= N
        \Phi{\left( \frac{x-\hat{x}_i}{\sigma_b} \right)}
        \biggr\rvert_{-L/2}^{L/2}
        \Phi{\left(\frac{y-\hat{y}_i}{\sigma_b} \right)}
        \biggr\rvert_{-L/2}^{L/2}
    \end{align*}
    where $\Phi(x)$ is the cumulative distribution function of the Normal
    distribution.

    \subsubsection{Network architecture and training}
    \label{sec:architecture}
    To predict the positions and detected photons of the simulated emitters
    described in Section \ref{sec:gauss_sim}, we trained a BNN consisting of the
    following layers (in order): a single convolutional layer with $3 \times 3$
    stride 1 kernels, 8 output features, and a leaky-ReLU activation; a
    fully-connected layer with 128 neurons and a leaky-ReLU activation; a
    fully-connected layer with 3 neurons; the Normal inverse sampling position
    node
    activation chain described in Section \ref{sec:icdf_activations}; and
    finally
    the Normal inverse sampling photon node activation chain described in
    \ref{sec:icdf_activations}.

    A total of 20 network replicates were trained as described in Section
    \ref{sec:training} from initial random seeds: 10 using \ut{} and 10 using
    \mc{}
    with $n=10$ samples to propagate through all nonlinear network layers.
    Notably, attempts to train this network architecture using \mc{} with $n=3$
    samples (the same number of samples used by \ut{}) resulted in exploding
    gradients, hence we used $n=10$ samples for all \mc{} networks in this
    experiment.

    \subsubsection{Results}
\begin{figure}[htbp!]
        \centering
        \includegraphics[width=\textwidth]{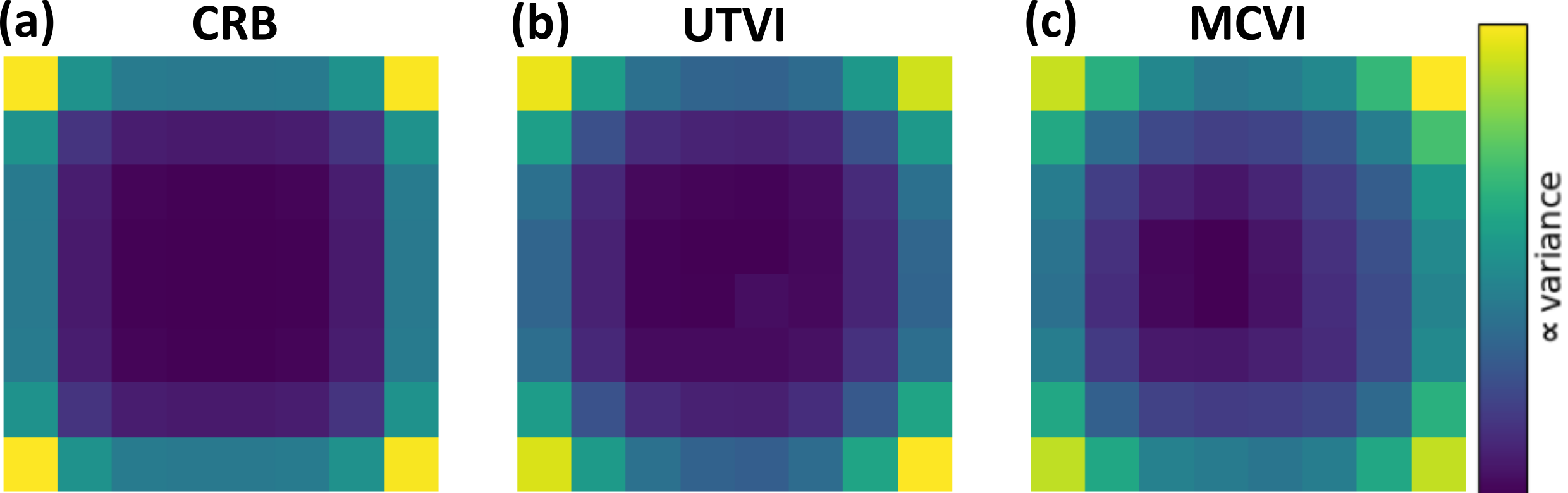}
        \caption{\textbf{\ut{} accurately predicts variance after
                nonlinear transformation.} (a) CRB (Eqn. \ref{eqn:crb}) of
                emitter position
            estimates as a
            function of position across $L \times L$ image. (b-c)
            Predictive variance of the position inferred by (b) \ut{} with 3
            sigma
            points and (c) \mc{}
            with 10 samples for
            the network architecture described in Section
            \ref{sec:architecture}.
            Images are scaled independently such that color is proportional to
            variance.}
        \label{fig:gauss_var}
    \end{figure}

    To compare the performance of \ut{} to \mc{} for the object localization
    task,
    we computed the predicted variance as a function of the true simulated
    emitter
    position within the image field of view.  This was done by simulating 1024
    randomly placed emitters within each pixel of the image, running inference
    using the 20 network replicates (10
    \ut{} and 10 \mc{} networks trained independently), and then averaging the
    position variances predicted by the networks within each pixel.  We compare
    the
    predicted variances to the Cram\'er-Rao bound (CRB) derived in
    \cite{ober2004}:
    \begin{equation}
        \label{eqn:crb}
        \mathrm{CRB}_i \approx \frac{\sigma_b^2}{N_i}
    \end{equation}
    where $N_i$ is the average
    number of photons collected by the sensor for each emitter simulated in
    pixel
    $\mathbf{r}_i$.  The results of this study are shown in Figure
    \ref{fig:gauss_var}.  Overall, \ut{} and \mc{} capture the basic trend
    (i.e.,
    higher variance near edges), though \ut{} more accurately captures the trend
    bounded by the CRB.

    To verify that our physics-informed activations defined in Section
    \ref{sec:icdf_activations} meaningfully improve network performance, we
    trained several object localization networks as in Section
    \ref{sec:architecture} but with fixed-length datasets defined prior to
    training.  As shown in Figure \ref{fig:activation}, our physics-informed
    activations improve network generalization as defined by the negative
    log-likelihood over an unseen validation set in the data-starved ($\leq
    2^9$ training pairs) regime.

    \section{Conclusions}
    In this work, we demonstrated that the unscented transform is an effective
    and
    efficient method for propagating the mean and variance through arbitrary
    nonlinear layers of BNNs, enabling few-sample variational inference of BNNs
    with an approximately 10X reduced cost as compared to Monte Carlo moment
    propagation.  We have shown that the unscented transform can be used to
    extend
    the analytical moment propagation methods of \cite{wu_deterministic_2019,
        hausmann_sampling-free_2020,schodt_2024} to networks where closed-form
        moment
    propagation rules are unavailable or intractable.  Additionally, we
    presented a
    novel nonlinear activation motivated by inverse transform sampling whose
    usage
    in BNNs is enabled or otherwise greatly simplified by our proposed
    few-sample
    inference scheme.  In summary, we have presented a novel approach to
    variational inference of BNNs that is simple to implement, computationally
    efficient, and extensible to arbitrary nonlinear BNN layers.

    %\section*{Acknowledgments}
    \begin{ack}
        We thank Dr. Alexander Wagner of Teledyne Scientific \& Imaging, LLC for
        useful discussions throughout this project and for providing feedback on
        the manuscript.
    \end{ack}

    \clearpage{}
    \bibliographystyle{unsrt}
    \bibliography{references/ref_file_primary.bib}

    \clearpage{}
    \appendix
    \section{Supplemental figures}

    \begin{figure}[htbp!]
        \centering
        \includegraphics[width=\textwidth]{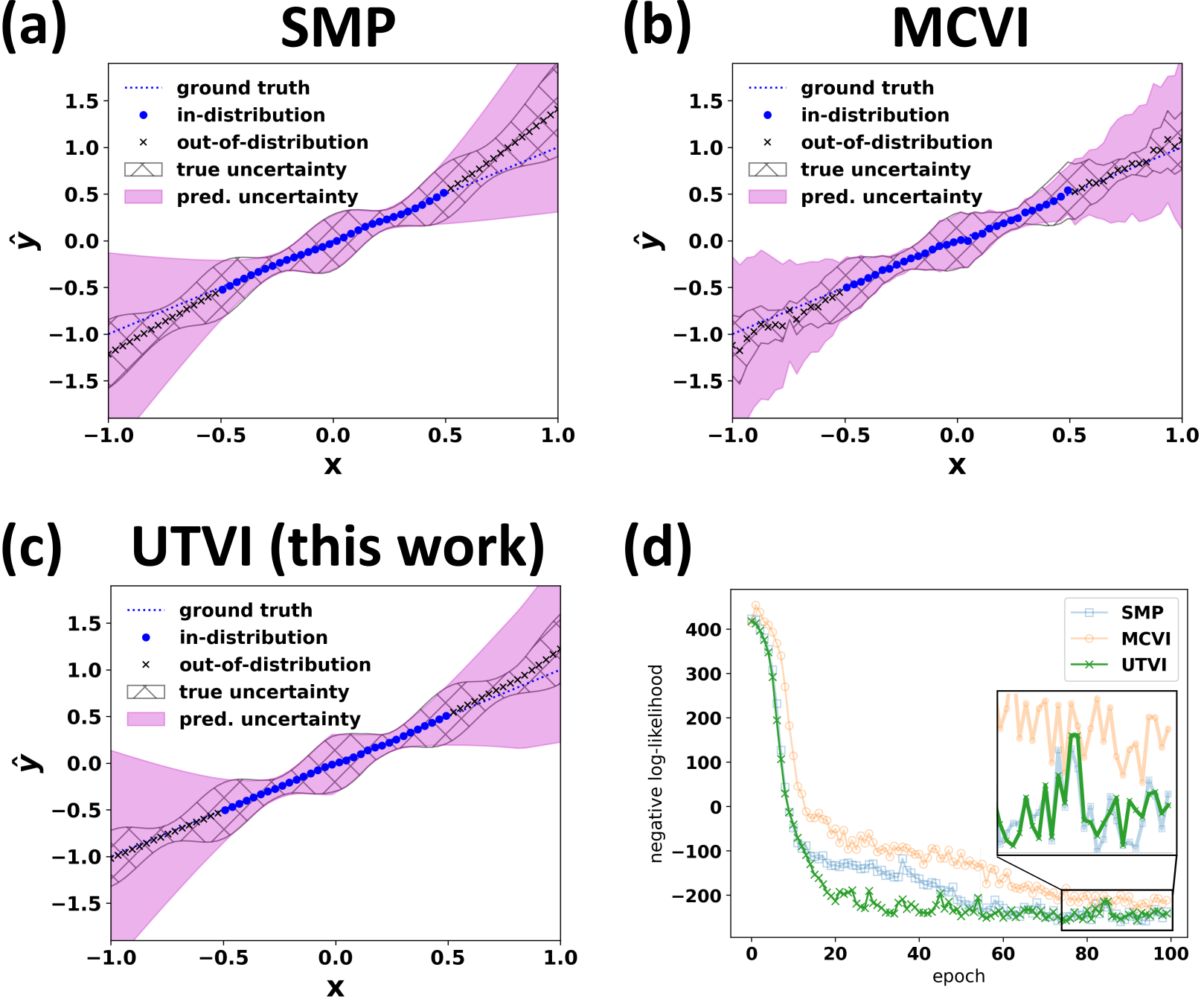}
        \caption{\textbf{\ut{} outperforms \mc{} and matches \smp{}.}  Example
            results from a single model (for each of \smp{}, \mc{}, and \ut{})
            for
            comparison to the averages over 10 models shown in Figure
            \ref{fig:fc_bnn}.  Note that \mc{} predictions are noisier than
            apparent from the model averaged results.}
        \label{fig:mc_3samples}
    \end{figure}

    \begin{figure}[htbp!]
        \centering
        \includegraphics[width=\textwidth]{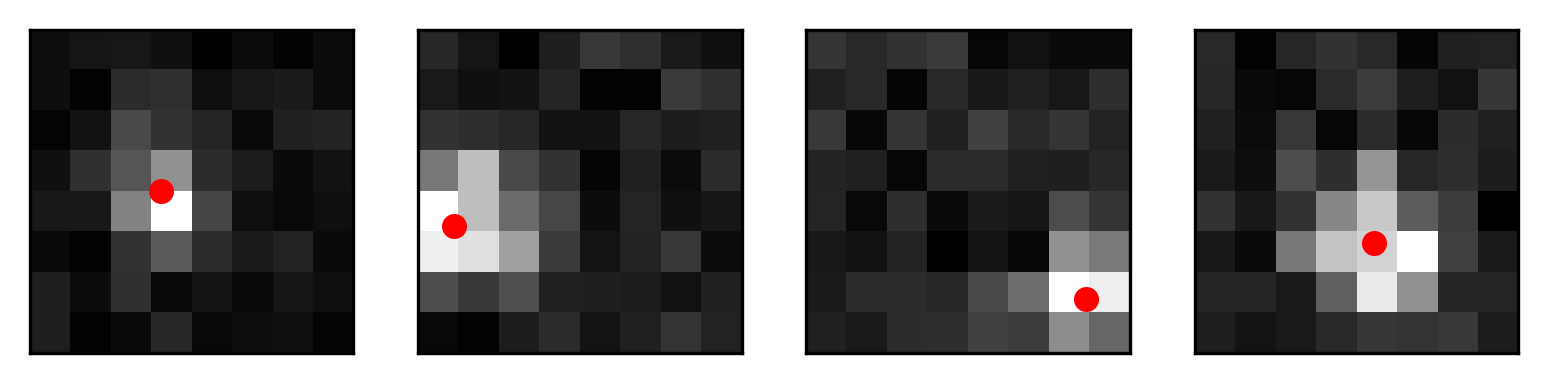}
        \caption{\textbf{Examples of simulated object localization data.} Random
            examples of data generated by the simulator described in Section
            \ref{sec:gauss_sim}.  Red dots are added for visualization to
            indicate
            the ground truth object position and are not included during network
            training or evaluation.  Images were rescaled independently to
            improve
            visualization.}
        \label{fig:ex_data}
    \end{figure}

    \begin{figure}[htbp!]
        \centering
        \includegraphics[width=\textwidth]{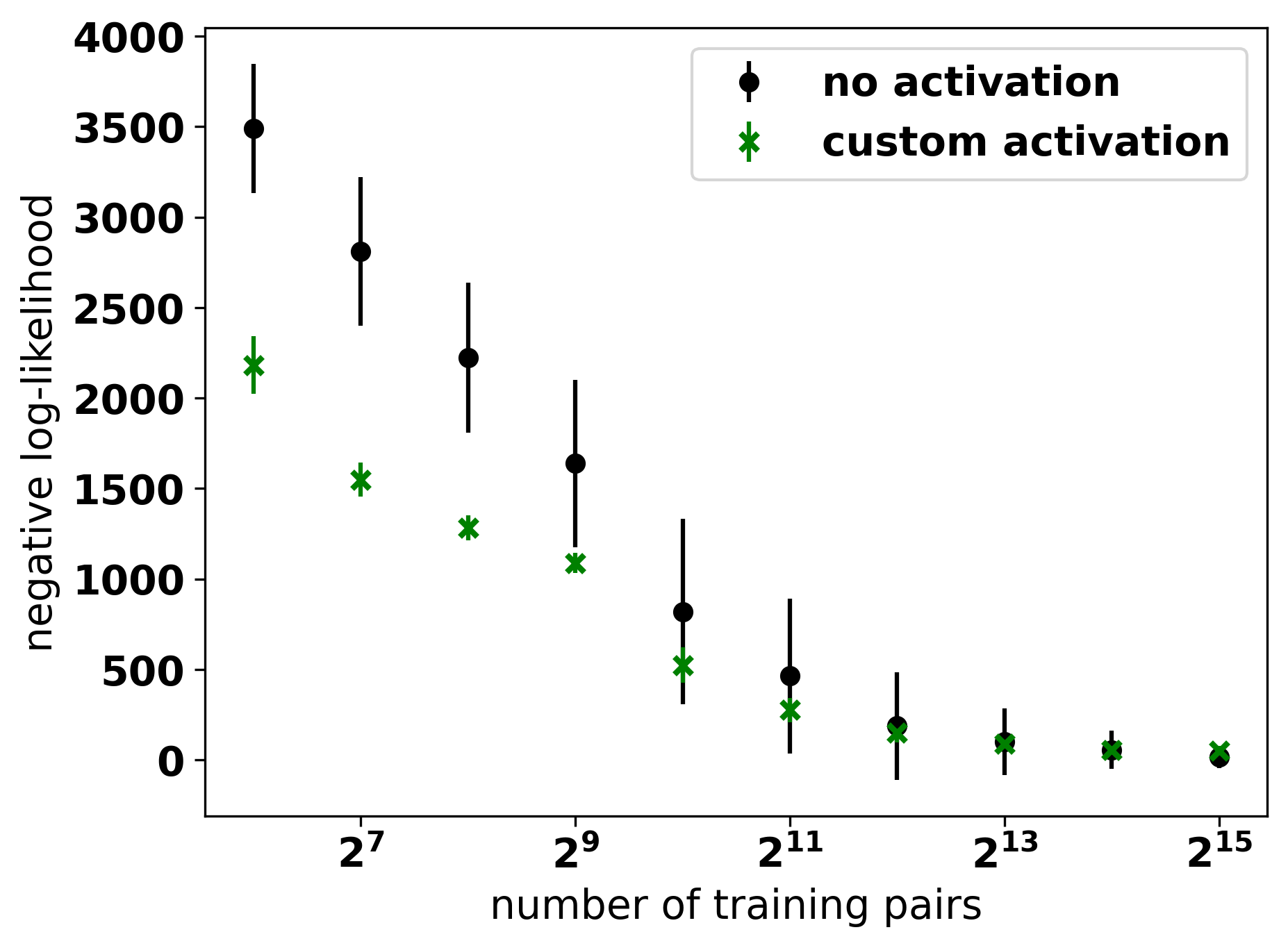}
        \caption{\textbf{Custom activations improve network generalization.}
        Minimum negative log-likelihood achieved during training over an
                unseen evaluation set when using no output node
                activation (black dots) versus the
                physics-informed activations described in Section
            \ref{sec:icdf_activations} (green x's).  Points represent the mean
            over 10 models trained independently on the same data starting
            from different random initializations, while error bars represent
            the standard deviation over the 10 models.}
        \label{fig:activation}
    \end{figure}

\end{document}